# A Deep Language-independent Network to analyze the impact of COVID-19 on the World via Sentiment Analysis

Ashima Yadav, Dinesh Kumar Vishwakarma, *Senior Member, IEEE*

**Abstract**—Towards the end of 2019, Wuhan experienced an outbreak of novel coronavirus, which soon spread all over the world, resulting in a deadly pandemic that infected millions of people around the globe. The government and public health agencies followed many strategies to counter the fatal virus. However, the virus severely affected the social and economic lives of the people. In this paper, we extract and study the opinion of people from the top five worst affected countries by the virus, namely USA, Brazil, India, Russia, and South Africa. We propose a deep language-independent Multilevel Attention-based Conv-BiGRU network (MACBiG-Net), which includes embedding layer, word-level encoded attention, and sentence-level encoded attention mechanism to extract the positive, negative, and neutral sentiments. The embedding layer encodes the sentence sequence into a real-valued vector. The word-level and sentence-level encoding is performed by a 1D Conv-BiGRU based mechanism, followed by word-level and sentence-level attention, respectively. We further develop a COVID-19 Sentiment Dataset by crawling the tweets from Twitter. Extensive experiments on our proposed dataset demonstrate the effectiveness of the proposed MACBiG-Net. Also, attention-weights visualization and in-depth results analysis shows that the proposed network has effectively captured the sentiments of the people.

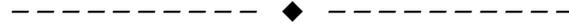

## 1 INTRODUCTION

THE outbreak of COVID-19 pandemic[1] (popularly referred to as coronavirus), caused by severe acute respiratory syndrome coronavirus 2 (SARS- CoV- 2), has badly hit the entire world. According to the World Health Organization (WHO)[2], this pandemic has affected 7,941,791 people with a death toll of 43,47,96 worldwide. Thus, the WHO declared a Public Health Emergency of International Concern on January 30, 2020. Nearly every country of the world has reported active cases, following which the WHO has advised them to follow strict COVID-19 mitigation guidelines like social distancing, contact tracing, non-essential business closure, quarantine, ban on travel and mass gatherings, etc. [1] [2]. This creates a strong need to assess the attitude and behavior of the people related to the COVID-19 pandemic.

Sentiment analysis is a popular branch in Natural Language Processing which extracts the opinions or views of the people embedded in user-generated data like online reviews, comments, and multimodal data [3] [4]. The field has shown tremendous success in areas like Business review analysis [5] [6], Financial market prediction [7], Recommendation system [8], Multimodal analysis [9]. In the medical domain, the sentiments reflect a patient's health status, which can be good (positive), bad (negative), or normal (neutral) [10]. They can be used for analyzing the drug and medication's effectiveness or confirming the presence or absence of disease [11]. Recently, deep learning models have demonstrated significant improvements in many text processing tasks [12] [13]. They avoid the traditional handcrafted feature engineering by automatically learning the semantic features from the training data itself.

Hence, in this paper, we propose a Multilevel Attention-based Conv-BiGRU Network (MACBiG-Net) for classifying the opinions of the people from the countries which were worse-affected by the COVID-19 pandemic. Firstly, the words are encoded into a real-valued vector with the help of the Glove embedding matrix. Then word-level encoding is performed by 1D CNN, where different convolution kernels help in learning the local features of the text, followed by BiGRU to capture the long term dependencies of a word in both left and right direction. After this, we apply the attention network to extract the most informative word. Finally, we follow the same combination of CNN- BiGRU for encoding the sentence vector and again using attention mechanism to compute the essential weights of the sentences for the entire document classification. In this way, MACBiG-Net can efficiently classify the text into positive, negative, and neutral sentiments. The significant contribution of our work is outlined as follows:

- We propose a deep Multilevel Attention-based Conv-BiGRU Network (MACBiG-Net) for classifying the opinions of the people from the top five worst affected countries by the coronavirus, namely USA, Brazil, India, Russia, and South Africa.
- We also develop the COVID-19 Sentiment dataset by

*Ashima Yadav and Dinesh Kumar Vishwakarma are with Biometric Research Laboratory, Department of Information Technology, Delhi Technological University, Delhi, India. E-mail: ashimayadavdtu@gmail.com, dvishwakarma@gmail.com.*

---

[1] https://www.who.int/dg/speeches/detail/who-director-general-s-opening-remarks-at-the-media-briefing-on-covid-19---11-march-2020

[2] https://www.who.int/docs/default-source/coronaviruse/situation-reports/20200815-covid-19-sitrep-208.pdf?sfvrsn=9dc4e959_2 (As on 15 August, 2020)

- crawling and downloading the tweets from the famous microblogging site, Twitter, and labelling them according to the sentiments belonging to the positive, negative, and neutral sentiment classes. We download and analyze the tweets of the people belonging to the USA, Brazil, India, Russia, and South Africa, starting from January 1, 2020, to June 7, 2020.
- The proposed MACBiG-Net is language-independent as it can classify the tweets belonging to multiple languages like Hindi, Japanese, Arabic, Spanish, Urdu, etc. effectively by encoding the words and sentences, and assigning larger weights to relevant contexts for the document classification.
- We conduct rigorous experiments on the proposed COVID-19 Sentiment dataset. Experimental results demonstrate that the proposed network show significant improvement over the popular baseline methods for sentiment classification in terms of Precision, Recall, F1 score, Accuracy, Receiver operating characteristic (ROC), and Area under the curve (AUC) evaluation metrics. We also visualize the attention weights of the network for getting in-depth knowledge about the attention mechanism.
- Finally, we draw several crucial observations by visualizing the country-wise sentiments of the people from January to June. We also analyze the popular positive and negative sentiment topics, words, and hashtags that were generated during the pandemic to see how the pandemic affected the lives of people all over the world. This analysis can serve as the feedback to the government agencies regarding the mitigation plans taken by them. Further, it may also guide the future planning of the public health agencies in case of any such outbreak.

The remainder of the paper is organized as follows: Section 2 discusses the Related Work for sentiment analysis in the health-care domain. Section 3 focuses on the proposed methodology by describing the proposed MACBiG-Net. Section 4 presents the experimental results by discussing quantitative and qualitative analysis. Section 5 concludes our paper and also focuses on future work.

## 2 RELATED WORK

The sentiment analysis is a vast field having ample applications, as discussed above. Since we could not find any previous work which had focused on sentiments expressed during the COVID-19 pandemic, hence we discuss prominent works related to sentiments in the health-care domain only. The facets of sentiments in medicine are broadly categorized into [10]: capturing patient's health status (good/bad/no change), diagnosing the certainty or severity of a disease, and the effectiveness of the treatment or drug.

The early work in this area focuses on in-built lexicons or machine learning-based approaches for extracting the sentiments. Yang et al. [14] proposed a framework based on the weighted scheme Latent Dirichlet Allocation (LDA) to analyze the health-related information regarding breast cancer. The medical topics were clustered together to perform sentiment analysis using AFINN lexicon. Each topic was associated with positive and negative sentiment terms. The framework can benefit other users to seek advice or take necessary precautions in case they experience similar symptoms. Sabra et al. [15] proposed a framework based on semantic extraction and sentiment assessment of the risk factors to predict the diagnosis of Venous thromboembolism disorder. The clinical narratives were parsed to extract the risk factors with the help of MetaMap. These risk factors were stored in a dictionary, and semantic-rich rules were derived from weighting the risk factors according to the severity. Finally, the polarity of the adjectives and adverbs were extracted with the help of SentiWordNet lexicon. The framework achieved 54.5% precision and 85.7% recall to identify patient's likelihood to develop VTE early and start the treatment as soon as possible.

Several clinical trials are necessary before releasing any drug to the market. Still, discovering all the side effects of the drugs is a challenging task. Moh et al. [16] developed an architecture to study the side effects of five crucial drugs by capturing the tweets from Twitter. The sentiment scores were extracted using several lexicons like SentiWordNet, AFINN, etc. Similarly, Jiménez-Zafra et al. [17] applied machine learning classifiers and lexicon-based approaches to compute the sentiment polarities for the physicians and the drugs. The reviews related to physicians contain informal language, while discussions about drugs contain the combination of colloquial language and specific terms like adverse effects or name of the drug. Korkontzelos et al. [11] identified the adverse drug reactions on social media platforms by integrating the sentiment feature. Experimental results show that sentiment features improved the F-score from 72.14% to 73.22%.

Recently, deep based approaches are being explored to extract the health-care sentiments expressed by the people. Limsopatham et al. [18] applied CNN and word embeddings on social media messages to medical concepts. Chen et al. [19] extracted the emotion features based on the LSTM network for screening the users on perinatal depression. The results show that the method improves the early screening process for perinatal depression. Grisstte et al. . [20] applied the Bi-LSTM model for identifying the adverse drug reactions from social media text. Talpada et al. [21] studied the impact of Telemedicine for heart attacks and epilepsy. They compared the lexical and semantic methods with the deep based methods. The lexical and semantic methods were based on VADER and TextBlob, whereas LSTM was employed for deep based classification. Experimental results signify that sentiments can be used to obtain the demographic distribution of Telemedicine for hearth attacks and epilepsy. Thus, sentiment in the context of medicine shows the impact of a disease or drug on the life of a person over a period of time. This can provide feedback to assess the effectiveness of the treatment or measures taken to combat a disease or a crucial event.

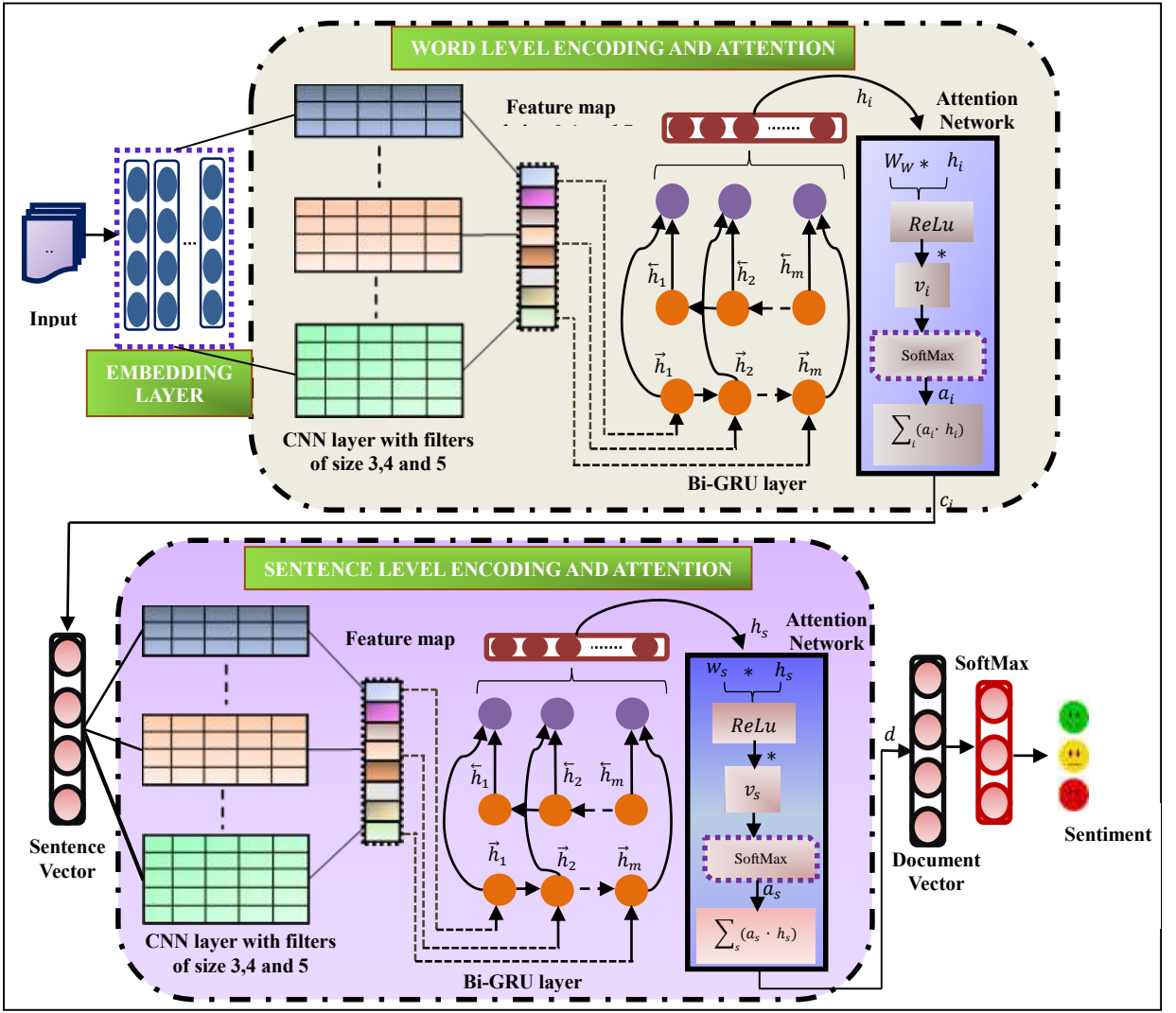

Fig. 1. Block diagram of the proposed Multilevel Attention-based Conv-BiGRU Network (MACBiG-Net)

## 3 PROPOSED METHODOLOGY

In this section, we explain the proposed framework for sentiment classification. It describes the data preprocessing process where we follow a two-step data cleaning approach, followed by the proposed Multilevel Attention-based Conv-BiGRU network (MACBiG-Net) to learn the semantic information of a sentence by encoding the essential words and sentences in a sequence

### 3.1 Data preprocessing

We scrapped the data from one of the most popular microblogging site Twitter for all the five countries based on their location. The complete data scraping, cleaning, and labeling process are discussed in Section 4. The final dataset comprises of people's views belonging to three output classes: positive, negative, and neutral. Since data processing is one of the most crucial tasks for handling the unstructured textual data, hence we meticulously designed a two-step data cleaning approach. The first approach is followed before translating the data into the English language so that the translation process is not affected by special characters or non-ASCII codes in the sentence. This includes removing hyperlink, special characters/symbols, retweet headers, twitter usernames, non-ASCII codes, NaN (Not a Number) character, and whitespaces. Finally, all the uppercase words are converted into lowercase by using the NumPy library functions.

In the second approach, the labeled dataset goes through stemming and lemmatization to reduce the word to its root form. Porter Stemmer and WordNet Lemmatizer are used for stemming and lemmatization, respectively. We also removed the stop words that frequently occur in the text and convey no meaningful information, followed by hashtags and punctuations removal. The cleaned data is then passed to the Multilevel Attention-based Conv-GRU network, as discussed in the subsequent section for the final classification.

### 3.2 Multilevel Attention-based Conv-BiGRU Network (MACBiG-Net)

The section describes the proposed MACBiG-Net, as shown in Fig. 1. The network includes three main steps: Embedding Layer, which encodes the input words into the low-dimensional vectors, Word-level Encoding and

Table I
Layer description of Step 2: Word-level Encoding and Attention

| Layer Name | Input Shape | Output Shape | # Parameters | Connected to |
|---|---|---|---|---|
| Input Layer | (None, 200) | (None, 200) | 0 | - |
| Embedding | (None, 200) | (None, 200, 100) | 1835200 | Input Layer |
| Conv1D_1 | (None, 200, 100) | (None, 198, 128) | 38528 | Embedding |
| Conv1D_2 | (None, 200, 100) | (None, 197, 128) | 51328 | Embedding |
| Conv1D_3 | (None, 200, 100) | (None, 196, 128) | 64128 | Embedding |
| MaxPooling1D_1 | (None, 198, 128) | (None, 66, 128) | 0 | Conv1D_1 |
| MaxPooling1D_2 | (None, 197, 128) | (None, 65, 128) | 0 | Conv1D_2 |
| MaxPooling1D_3 | (None, 196, 128) | (None, 65, 128) | 0 | Conv1D_3 |
| Concatenate | (None, 66, 128), (None, 65, 128), (None, 65, 128) | (None, 196, 128) | 0 | MaxPooling1D_1, MaxPooling1D_2, MaxPooling1D_3 |
| MaxPooling1D_4 | (None, 196, 128) | (None, 65, 128) | 0 | Concatenate |
| Bidirectional_GRU | (None, 65, 128) | (None, 65, 200) | 183200 | MaxPooling1D_4 |
| TimeDistributed (Dense) | (None, 65, 200) | (None, 65, 100) | 20100 | Bidirectional_GRU |
| Hierarchical_attention_Network | (None, 65, 100) | (None, 100) | 10200 | TimeDistributed (Dense) |

Attention, in which the words of a sentence are encoded by 1D Conv-BiGRU based representation to obtain the sentence representation, followed by word-level attention which computes the essential weights for the sentence vector, and Sentence level encoding and attention, where the sentences of a document are again encoded by 1D Conv-BiGRU based representation to obtain the document representation, followed by sentence-level attention to get the document vector for the classification.

### 3.2.1 Embedding Layer

Each word in the sequence of the sentence is converted into a real-valued vector. Formally, let $\{r_1, r_2, \ldots, r_n\}$ be the sequence of $n$ words in a sentence. We first embed each word $w_i$ into a real-valued vector through a word embedding matrix $E \in R^{d_r*|V|}$, where $d_r$ is the word embedding size, and $V$ is the vocabulary size. We use the Glove[3] embedding matrix to get the final sequence of vector $\{e_1, e_2, \ldots, e_n\}$, which serves as an input for the next layer.

### 3.2.2 Word-level Encoding and attention

Our proposed architecture is inspired by [22], which intends to capture the structure of words from the sentence and the structure of sentences from a document. The word encoding is performed by 1D convolution, followed by the Bi-GRU mechanism. We first describe the convolution process as follows:

The local features are extracted by using a 1D CNN network with different kernel sizes to generate a feature map. The different convolution kernels help in learning the various local characteristics of the text. In general, for $e_i^{th}$ embedding corresponding to $r_i^{th}$ word, the concatenation operation ⊕ is expressed as follows:

$$e_{1:n} = e_1 \oplus e_2 \oplus \ldots \oplus e_n \qquad (1)$$

The convolution operation is applied on the window of $j$ words to generate a new feature value $m_i$ as shown below:

$$m_i = f(W \cdot e_{i:i+j-1} + b) \qquad (2)$$

Where, $b$ = bias and $f$ = Rectified Linear unit activation function (ReLu). This filter is applied to different possible window of words in a sentence which generates the following feature map:

$$M = [m_1, m_2, \ldots, m_{n-j+1}] \qquad (3)$$

We apply the filter of sizes 3, 4, and 5 to obtain multiple features followed by max-pooling layer to capture the maximum feature value. The concatenated output of all three filter sizes is passed into the BiGRU based layer, which contains the hidden states to store crucial information. The bidirectional nature of GRU is based on the fact that words in a sentence are not only related to the previous words but also to the following words. We have chosen GRU over LSTM due to its efficiency in training, which makes it computationally cheaper than LSTM. The forward GRU with $\vec{h}_i = \overrightarrow{GRU}(e_i, \vec{h}_{i-1})$ is concatenated with backward GRU, $\overleftarrow{h}_i = \overleftarrow{GRU}(e_i, \overleftarrow{h}_{i+1})$ to obtain the output of BiGRU at step $w$ as $h_i = [\vec{h}_i, \overleftarrow{h}_i]$. Thus, the BiGRU layer fetches the contextual information of every sentence centered around the word $w_i$. This makes sense as the orientation of sentiment needs to consider the past and future context information in a sentence. Thus, the word is encoded by the Conv-BiGRU block by storing the local and contextual features.

The attention mechanism is based on the idea that when users read a document word by word in each sentence, then they would pay attention to the most informative word or sentence. Hence, we model both the word level and sentence level attention differently. In word-level attention, the word annotation $h_i$ is fed into the network to get its vector representation $v_i$. We apply the *ReLu* activation function over *tanh* as specified in [22], as *ReLu* converges quickly and results in cheaper computation. This is described in the (4):

$$v_i = ReLu(W_w \cdot h_i + b_w) \qquad (4)$$

---

[3] https://nlp.stanford.edu/projects/glove/

Table II
Layer description of Step 3: Sentence-level Encoding and Attention

| Layer Name | Input Shape | Output Shape | # Parameters | Connected to |
|---|---|---|---|---|
| Input Layer | (None, 15, 200) | (None, 15, 200) | 0 | - |
| TimeDistributed (Model) | (None, 15, 200) | (None, 15, 100) | 2202684 | Input Layer |
| Conv1D_1 | (None, 15, 100) | (None, 13, 128) | 38528 | TimeDistributed (Model) |
| Conv1D_2 | (None, 15, 100) | (None, 12, 128) | 51328 | TimeDistributed (Model) |
| Conv1D_3 | (None, 15, 100) | (None, 11, 128) | 64128 | TimeDistributed (Model) |
| MaxPooling1D_1 | (None, 13, 128) | (None, 4, 128) | 0 | Conv1D_1 |
| MaxPooling1D_2 | (None, 12, 128) | (None, 4, 128) | 0 | Conv1D_2 |
| MaxPooling1D_3 | (None, 11, 128) | (None, 3, 128) | 0 | Conv1D_3 |
| Concatenate | (None, 4, 128), (None, 4, 128), (None, 3, 128) | (None, 11, 128) | 0 | MaxPooling1D_1, MaxPooling1D_2, MaxPooling1D_3 |
| MaxPooling1D_4 | (None, 11, 128) | (None, 3, 128) | 0 | Concatenate |
| Bidirectional_GRU | (None, 3, 128) | (None, 3, 200) | 183200 | MaxPooling1D_4 |
| TimeDistributed (Dense) | (None, 3, 200) | (None, 3, 100) | 20100 | Bidirectional_GRU |
| Hierarchical_attention_Network | (None, 3, 100) | (None, 100) | 10200 | TimeDistributed (Dense) |
| Dropout | (None, 100) | (None, 100) | 0 | Hierarchical_attention_Network |
| Dense | (None, 100) | (None, 3) | 303 | Dropout |

The resulting annotation $v_i$ is multiplied (dot product) by the word context vector $v_{wt}$ which is learned during the training. The softmax function is applied to obtain attention weights $a_i$ as shown in (5) below:

$$a_i = \frac{\exp(v_{wt} \cdot v_i)}{\sum_i \exp(v_{wt} \cdot v_i)} \quad (5)$$

Finally, the attention weights $a_i$ is concatenated with the word annotations to obtain the sentence vector $c_i$ as described in (6) below:

$$c_i = \sum_i (a_i \cdot h_i) \quad (6)$$

The layer details of the word-level encoding and attention step are shown in Table I.

### 3.2.3 Sentence level encoding and attention

As discussed, selecting important sentences in the document becomes another crucial task for the classification. Hence, we follow the same procedure for generating the document representation of the sentences and, as discussed in Section 3.2.2, by encoding the sentences with relevant context and computing the crucial weights of these contexts for the document classification.

The sentence vector $c_i$ obtained in Section 3.2.2 is passed into the 1D CNN layer and Bi-GRU layer to encode the sentence and obtain information $h_s$ of the neighboring sentences centered around sentence $s$. The sentence context vector $v_{st}$ is used to compute the sentence level attention $a_s$, which further yields the document vector $d$ that summarizes all the information of the sentences in a document. Mathematically, this is represented as below in (7)-(8).

$$v_s = ReLu \ (W_s \cdot h_s + b_s) \quad (7)$$
$$a_s = \frac{\exp(v_{st} \cdot v_s)}{\sum_s \exp(v_{st} \cdot v_s)} \quad (8)$$
$$d = \sum_s (a_s \cdot h_s) \quad (9)$$

The layer details of the Sentence level encoding and attention step are shown in Table II. The learned vector $d$ of [1* 100] dimension is passed to the dropout layer with a 0.5 rate to handle the overfitting problem in deep neural network architectures. This is followed by a dense layer and softmax activation function. The entire network is trained end to end for obtaining the best weights by monitoring the validation accuracy of the model. We use categorical cross-entropy loss, which is defined in (10) as follows:

$$Loss = -\sum_{j=1}^{3}(y_j \cdot \log \hat{y}_j) \quad (10)$$

Where, $y_j$ is the target value corresponding to the model output $\hat{y}_j$ for the $j^{th}$ sample. We have used the L2-regularization parameter, which penalizes the larger weights in the network for avoiding the high variance problem. This makes the objective function of the network as follows:

$$Cost \ function = Loss + (\lambda/2m * \sum ||w||^2) \quad (11)$$

Here, $\lambda$ is the regularization parameter and one of the hyperparameters of the network, which is tuned according to the validation set. Our model achieves optimized results with $\lambda = 0.001$.

## 4 EXPERIMENTAL ANALYSIS

This section discusses the data collection and labeling process in Section 4.1, implementation details in Section 4.2. We evaluate the effectiveness of the network in Section 4.3 and perform baseline comparison in Section 4.4. Finally, we visualize our classification results and extract the sentiment topics in Section 4.5.

### 4.1 Dataset Collection and Labeling

In order to analyze the impact of the novel coronavirus, we extracted the Tweets from January 1, 2020 to June 7, 2020 of top five most affected countries, namely, USA (52,03,206 cases), Brazil (32,24,876 cases), India (25,26,192 cases), Russia (9,17,884 cases), and South Africa (5,79,140 cases) as per data available from WHO[4]. We used some popular hashtags like #coronavirus, #covid19, and #COVID_19. The scrapped tweets were preprocessed as discussed in Section 3.1 and translated into English by the Google translator. Some sample tweets in a different language and their English translation is shown in Fig. 2.

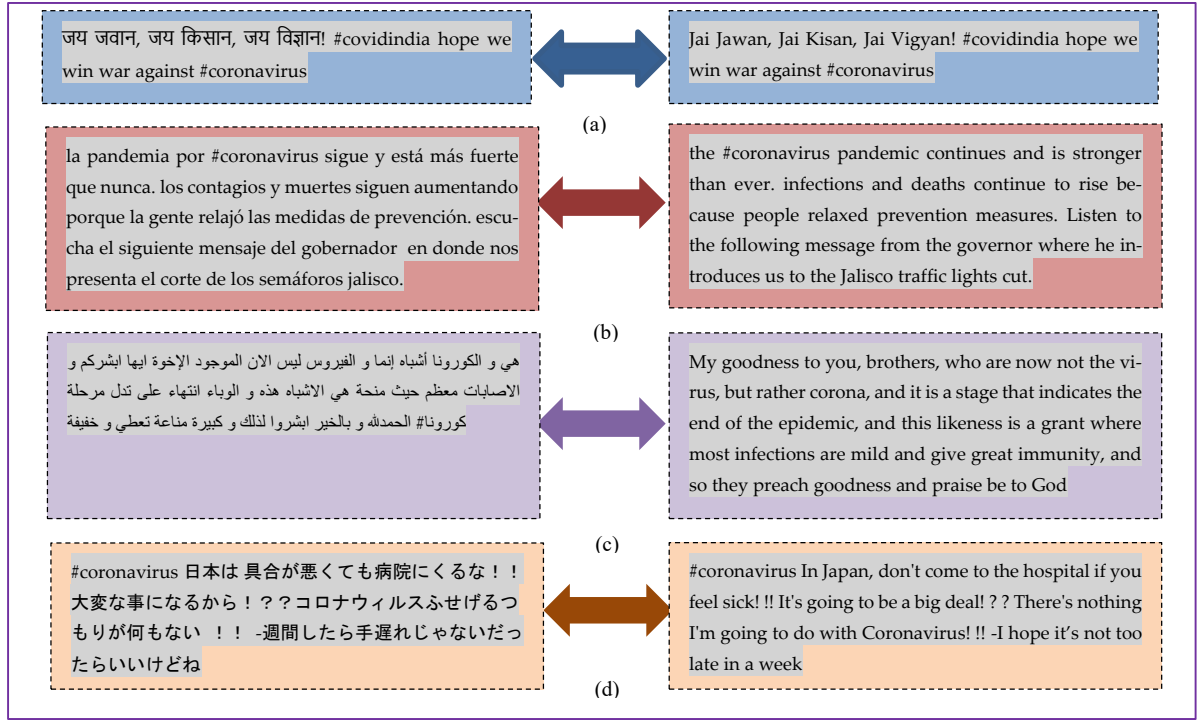

**Fig. 2.** Sample tweets in different languages from COVID-19 Sentiment Dataset (a) Hindi to English (b) Spanish to English (c) Arabic to English (d) Japanese to English.

**Table III**
**Country-wise details of COVID -19 Sentiment Dataset**

| Country | Sentiment | | | Total |
|---|---|---|---|---|
| | Positive | Negative | Neutral | |
| USA | 383 | 321 | 431 | 1135 |
| Brazil | 267 | 318 | 444 | 1029 |
| India | 302 | 164 | 299 | 765 |
| Russia | 147 | 168 | 249 | 564 |
| South Africa | 200 | 193 | 232 | 625 |
| **Total** | **1299** | **1164** | **1655** | **4118** |

We used the Textblob[5] library to weakly label the processed tweets with the sentiment score of [-1, 1]. The tweets with +1 sentiment scores were labeled as positive, tweets with -1 score were labeled as negative, and finally, tweets with 0 sentiment scores were labeled as neutral. However, we observe that in many cases, the Textblob library fails to classify the data into correct sentiment labels. E.g., Consider the following tweet, "*My dad tested positive for the corona*" was labeled as positive sentiment, though it belongs to the negative sentiment category. Hence, the shortlisted tweets were again manually annotated (strong labels) by three humans with positive (+1), negative (-1), and neutral (0) sentiment category. Finally, 4118 tweets were annotated with an average value of Cohen Kappa inter-annotator agreement as 0.85. The country-wise dataset details corresponding to each sentiment category are shown in Table III.

### 4.2 Implementation Details

The proposed MACBiG-Net is implemented on Python 3 using the popular Keras framework. The experiments were performed on Windows 10 machine with 128GB RAM using NVIDIA Titan RTX GPUs. The embedding dimension is 100, maximum length of all sequences is 200, and Punkt sentence Tokenizer is used for dividing the text into a list of sentences. Adam optimizer with learning rate = 0.0001, default parameters $\beta_1 = 0.9$ and $\beta_2 = 0.999$ is set for training. The entire network is trained end to end with 64 batches for 100 epochs. Dropout and Regularization are used for avoiding overfitting. We have used a ten-fold cross-validation testing strategy with 80% data for training, 5% data for validation, and 15% data for testing. The validation accuracy is monitored during training the network. The model achieving the highest validation accuracy is selected for the testing phase.

### 4.3 Classification Results and Observations

We compare the sentiment prediction performance of the MACBiG-Net in terms of precision, recall, F1 score, and accuracy corresponding to different sentiment categories, as shown in Fig. 3 (a). Since we used a ten-fold cross-validation strategy, the final results are reported by averaging the results across each of the test fold.

The accuracy of positive, negative, and neutral sentiment category is 79.9%, 80.2%, and 83.6%, respectively.

---

[4] https://www.who.int/docs/default-source/coronaviruse/situation-reports/20200815-covid-19-sitrep-208.pdf?sfvrsn=9dc4e959_2 (As on 15 August, 2020)
[5] https://textblob.readthedocs.io/en/dev/

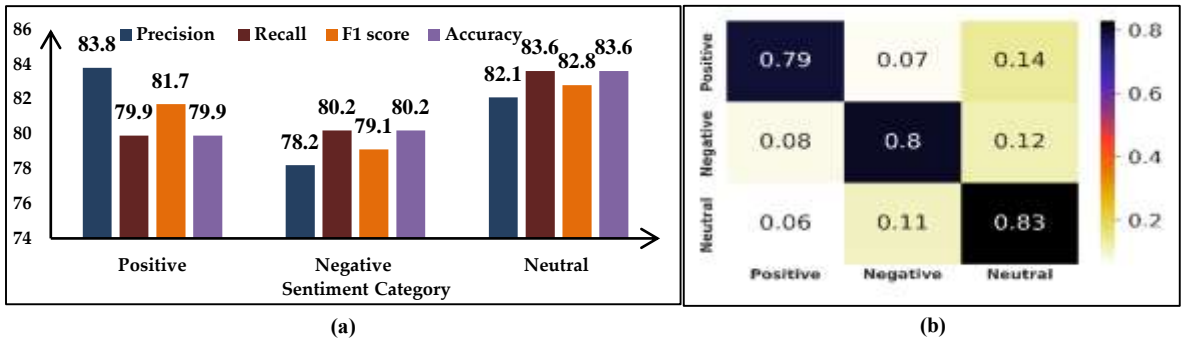

Fig. 3. (a) Sentiment classification results on COVID-19 Sentiment Dataset (%) (b) Confusion Matrix.

The average accuracy of the network is 81.5%. Further, the confusion matrix in Fig. 3 (b) gives an in-depth analysis of the classification results. It can be seen that a major amount of misclassification occurs when a positive sample is incorrectly classified as neutral. Yet, we can see that the network can discriminate well across each of the sentiment categories. These results clearly explain the effectiveness of MACBiG-Net.

We also provide accuracy and loss curves for training and validation set in Fig. 4 (a) and (b), respectively. As we can see, the training accuracy is 100%, which means the model is trained completely, and the validation accuracy shows how effectively the model can perform on the unseen samples. Similarly, decreasing loss curves with each epoch confirm the adequate learning of the model and validate its internal classification performance.

The foremost aim of our work is to analyze the sentiments of people expressed globally on the COVID-19 pandemic. This can be visualized in Fig. 5, where we plot the positive (+1), negative (-1), and neutral (0) sentiments of people over time located in the top five worst affected countries by the pandemic. As discussed, we retrieved the tweets from January 1, 2020 to June 7, 2020. However, in many countries, sentiments were generated in the month of late February or March. This may be because the first confirmed case in that country started in late February (Eg: *The virus was first confirmed in Brazil on February 25, 2020* [6]), or no specific sentiments were generated during January and February. A five-day window represents the time component in Fig. 5 along the x-axis. This means that each bar represents the sentiment expressed in these five days. However, each day, several sentiments were being generated. Hence, we took the sentiment that was voiced the maximum number of times by the people on any particular day as the final sentiment for that day. We summarize our observations as follows:

- In the USA, significant sentiments emerged from March onwards. This may be because Trump declared a national emergency[7] on March 13, and by the end of March, the cases were reported in all 50 US States[8]. In early March, positive sentiment was recorded. However, by the end of March till mid-April, negative sentiment was prevalent. Similarly, by the end of May, we observed mostly negative sentiments. One of the primary reasons was the breaking news of George Floyd, which created a state of panic and terror in the USA from May 25 onwards. People started tweeting that COVID-19 will continue to spread as citizens were protesting and moving in groups, without following any preventive measures. Overall, it was observed that 43.6% of sentiments were neutral, 37.9% were positive, and 18.3% were negative.

- In India, positive sentiments were observed majorly in March and April. This may be because people were hopeful in the initial stages to overcome the virus. Further, Prime Minister (PM) Narender Singh Modi came up with some ideas to boost the support of all the frontline workers. On March 22, the PM requested every Indian to clap their hands or ring some bells from their house to boost the morale of frontline workers. Similarly, on April 5, the PM urged everyone to light candles or *diyas* for nine minutes as a mark to fight against the deadly virus. Several people followed these tasks and happily enjoyed them. Towards the end of May, we see the people expressed negative sentiments. Since, by the end of May, a large number of active cases were being reported each day. Overall, 50.6% of positive sentiments, 28.3% of neutral sentiments, and 20.9% of negative sentiments were observed in India.

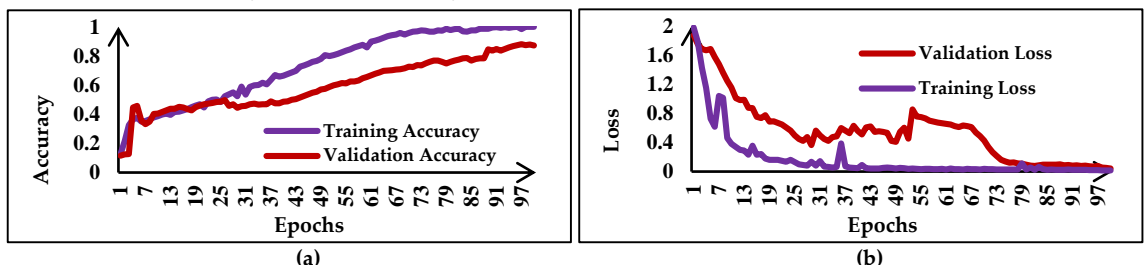

Fig. 4. Tracing internal performance of MACBiG-Net with (a) Accuracy curve for train and validation set (b) Loss curves for train and validation set.

---

[6] https://www.gov.br/saude/pt-br
[7] https://edition.cnn.com/2020/03/13/politics/donald-trump-emergency/index.html
[8] https://archive.vn/20200428073138/https://banningca.gov/DocumentCenter/View/7139/CDC_COVID19-Weekly-Key-Messages_03292020_FINAL

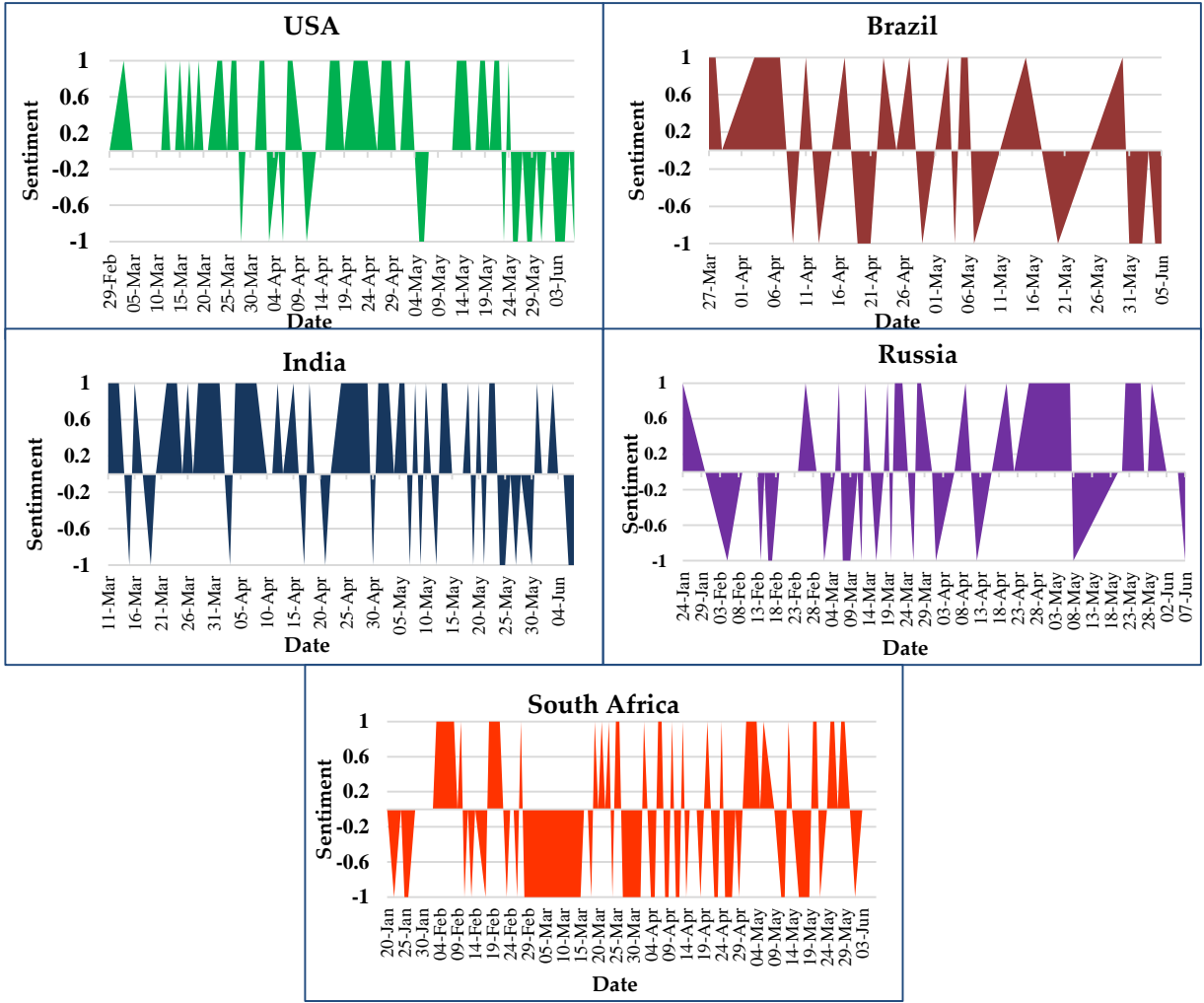

**Fig. 5. Visualizing the sentiments of people over time in top five worst affected countries by COVID-19 pandemic.**

- Russia reported its first positive case on January 31, 2020. People have majorly shown neutral and negative sentiments during the mid of February. However, positive sentiments were expressed in April. Overall, 39.2% of neutral sentiment, 33.9% of positive sentiment, and 26.7% of negative sentiment were observed in Russia.
- In South Africa, majorly negative sentiments were being observed from late February till mid of March. People finally realized that the deadly virus had hit their country as the number of cases started rising sharply. However, in April, we can see some positive sentiments as citizens started motivating everyone by asking them to wear masks, read books, involve in daily exercise, and posting positive ideas by their former President Nelson Mandela. Overall, South Africa expressed 39.6% of negative sentiments, 32.6% of positive sentiments, and 27.7% of neutral sentiments.
- Finally, we conclude that overall, 38.3% of positive sentiments, 32.6% of neutral sentiment, and 28.3% of negative sentiments were expressed by these top five affected countries.

### 4.4 Baseline Comparison

We use several popular baseline methods for comparison with the proposed MACBiG-Net. The details are discussed as follows:

**Table IV**
**Comparative Results of different methods on COVID-19 Sentiment Dataset (%)**

| Method | Accuracy | Precision | Recall | F1 score |
|---|---|---|---|---|
| LSTM [23] | 72.6 | 73.3 | 73.6 | 73.4 |
| Bi-LSTM [24] | 73.2 | 74.9 | 72.0 | 73.4 |
| CNN [25] | 75.8 | 73.1 | 75.2 | 74.1 |
| CNN-RNN [26] | 77.7 | 78.6 | 74.2 | 76.3 |
| HAN [22] | 79.4 | 77.4 | 78.9 | 78.1 |
| Self-attention-based LSTM [27] | 70.9 | 68.2 | 66.7 | 67.4 |
| RNN-Capsule [28] | 71.5 | 68.5 | 70.8 | 69.6 |
| BERT [29] | 76.6 | 75.2 | 74.8 | 74.9 |
| **MACBiG-Net (Ours)** | **81.5** | **81.3** | **81.2** | **81.2** |

- **LSTM** [23]**:** The classical long-short memory network is applied. Word vectors are initialized by Glove. The hidden layer size is 300 units. Adam optimizer with learning rate of 0.001 is used.
- **Bi-LSTM** [24]**:** This represents standard bi-directional LSTM. We follow the same settings as described above for LSTM.
- **CNN** [25]**:** Convolutional neural network with multiple filters of size [3,4,5], each having 100 filters is applied. Adam optimizer is used with a learning rate of 0.001.
- **CNN-RNN** [26]**:** We successively stack the CNN layer with max-pooling layer followed by RNN units. For RNN, we experimented with LSTM and GRU variants. The GRU based integration shows the highest performance.
- **HAN** [22]**:** The hierarchical attention network uses two LSTM layer followed by attention mechanism at word-level and sentence-level for composing the final text representation. The HAN is the simplest form of our proposed approach as it does not encounter localized features.
- **Self-attention based LSTM** [27]**:** The self-attention mechanism is used with Bi-LSTM, which learns the contribution of each hidden state by providing a set of summarized weight vectors for each hidden state. A learning rate of 0.06 is used.
- **RNN-Capsule** [28]**:** We apply capsule networks where the input instance representation is taken from the LSTM unit. The dimension of hidden vectors is set to 256. The attention mechanism is used to construct the capsule representation inside another capsule. All word vectors are initialized with Glove. Adam optimizer is used with a learning rate of 0.001.
- **BERT** [29]: We use a pre-trained Bidirectional Encoder Representation from Transformers (BERT) model with the help of an online released TensorFlow[9] library. The transformer blocks are set to 12, hidden size to 768, and self-attention head to 12.

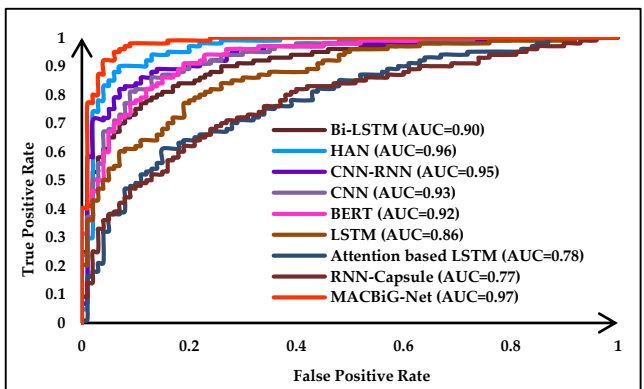

**Fig. 6. Comparison of MACBiG-Net with baseline methods in terms of ROC curve and Area under the curve (AUC).**

As evident from Table IV, the proposed network has outperformed the baseline methods and recent works in terms of sentiment classification by achieving 81.5% average accuracy. However, the HAN [22] and CNN-RNN [26] provides good enough results for the classification. This serves as the primary motivation of our proposed MAC-BiG-Net, which inherits the basic idea of both the methods and achieves nearly 2% and 4% higher accuracy compared to [22] and [26], respectively. This can also be visualized from the ROC curves and area under the curve (AUC), shown in Fig. 6, where we plot the ROC curves for the baseline methods. The AUC values help in comparing the ROC curves in a better way. From the AUC values, it is clear that MACBiG-Net has consistently outperformed all the previous methods for sentiment classification.

### 4.5 Qualitative Analysis

In this section, qualitative analysis is performed to visualize the word-level attention and extract the positive and negative sentiment topics that were generated during the pandemic.

#### 4.5.1 Visualizing Word-level Attention

In this section, qualitative analysis is performed on MAC-BiG-Net, to investigate the role of word-level attention for the sentiment classification. We randomly picked four tweets from our dataset, two of them were positive, and two were negative. Thus, in Fig. 7, (a) – (b) belongs to the positive sentiment class (shown in Red color), and (c) – (d) belongs to the negative class (shown in Blue color). The color signifies the attention weights of each word. The darker the color in the respective category, the higher amount of attention is given to the word.

We can see that attention focuses on the prominent part of the sentence, by giving darker color to the most important word. E.g., In review sentences (a) - (b), positive sentiment is conveyed by the crucial words like *'brilliant'*, *'happy'*, *'healthy'*, *'safety'*, *'good'*, which have received more score by the attention mechanism for better sentiment prediction. Similarly, in reviews (c)-(d), negative sentiment is reflected by words like *'scary'*, *'pandemic'*, *'death'*, which again have received more attention score. The words which convey no informative meaning (like stop words), were given very low weights by highlighting them with light colors. This shows that the attention weights focus on those words which have more relation with the output sentiment category. Thus, the visualization shows how the attention weights are changing along with the words in a sentence.

#### 4.5.2 Visualizing positive and negative sentiment topics

To get an in-depth analysis on the opinion of the people, we apply LDA (Latent Dirichlet Allocation) based topic modeling to determine the set of words (topics) that contribute to generate the positive and negative sentiments. This will help in finding the hidden thematic structure of the document and drawing essential conclusions from them. We use LDAvis [30] tool for visualization, which further helps in finding the prevalent topics in a document, relation between the topics, and the conditional distribution of a given term over the set of topics. The optimal number of topics was determined by building many LDA models with a different number of topics ($k$) and selecting the one with the highest coherence score [31]. This gives us k =10. The layout of LDAvis is shown in Fig. 8 (a)-(b) for positive and negative sentiment labels, respectively.

---
[9] https://github.com/tensorflow/tensor2tensor

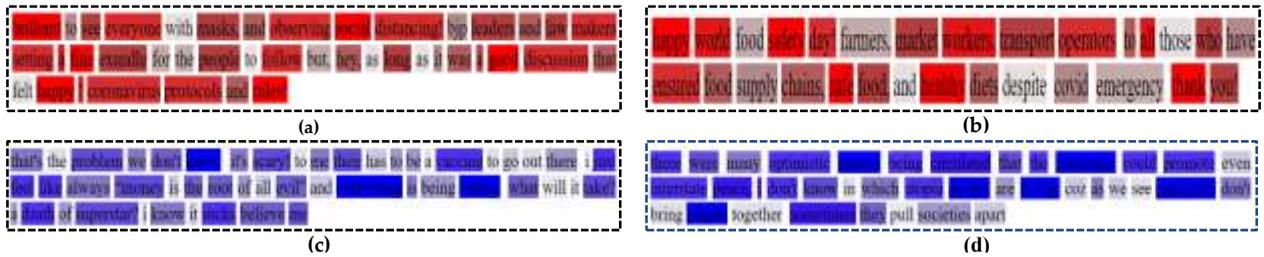

Fig. 7. Visualization the word-level attention weights from proposed dataset: (a)-(b) Positive Sentiment (c)-(d) Negative Sentiment.

If we consider (a), the entire layout is divided into two panels: the left panel shows the ten topics in the form of circles where the similar topics are plotted close to each other, the right panel shows the bargraphs which represent the terms or words that are most useful for understanding the currently selected topic (from the left panel). The length of the gray bars depicts the entire document-wise frequencies of each term, and the red bars depict the topic-wise frequencies of each term. Thus, for a given topic, we can compare the length of the red and gray bars to understand the relevance of the topic. Further, selecting a term on the right panel tells the conditional distribution of the selected term over the topics (on the left panel). In this way, topic modeling helps in explaining the underlying information in a document. Hence, we summarize our observations as follows:

- The prevalent positive words related to COVID-19 on twitter are: 'good', 'stay safe', 'happy', 'best', 'thank you', 'support', 'doctors', 'mask', 'together', 'proud' and the popular negative words linked to COVID-19 on twitter are: 'virus', 'deaths', 'bad', 'trump', 'help', 'lost', 'fake', 'died', 'outbreak', 'infected', 'hate', 'worst', 'crisis', 'fight'.
- Many positive tweets were directed towards the frontline workers where users were motivating and boosting the morale of the health care workers by appreciating them, donating them masks, dedicating some songs, providing free accommodation to them, and donating money to restaurants for distributing meals to essential workers.
- In some countries like Brazil, health care workers celebrate the discharge of the first COVID-19 patient from the hospital. In India, many citizens were delighted to see everyone unified with candles or flashlights in their hands from their balconies.
- Finally, positive sentiments were directed towards maintaining a healthy diet, reuniting with families after travel restrictions were lifted, boosting the immune system, asking everyone to stay safe, binge-watching television series, encouraging everyone to take precautions, wishing and motivating the users on special occasions by posting positive messages on occasions like Birthdays, Mothers Day, Earth Day, World Bicycle Day, Memorial Day (celebrated in the US), World Labor Day, International Nurses Day, and staying together to beat the deadly virus.
- During the pandemic, several social-economic factors also contributed to generate negative sentiments in people all around the world. In the US, the negative tweets were mainly focused on the racial injustice related to the death of George Floyd, which resulted in massive protests and violence even during the time of the pandemic. People fear that these mass demonstrations may occur in an uptick in the COVID-19 cases.
- In India, Mumbai (one of the worst-hit cities by COVID-19), was severely affected by cyclone winds. In West Bengal, Cyclone Amphan caused extensive damage. The poor, jobless migrants went traveling back to their home by walking thousands of kilometers due to lockdown restrictions in India.
- In general, many people also lost their jobs. The unemployment rate was increased during the pandemic. People were bored and anxious during home-quarantine. The upsurge in infected cases also created fear and stress among people. Further, lots of fake news[10] was generated in social media, creating panic about the disease.
- The popular positive hashtags are: #salutetocoronafighters, #inthistogether, #lovemycountry, #withme, #thankyouindia, #indiafightscorona, #flattenthecurve, #fightthevirus, and #frontlineheroes and negative hashtags are: #DictatorTrump, #BlackLivesMatter, #GeorgeFloyd, #AmericaInCrisis, #Riot2020, #migrantlivesmatter, #pandemia, #chinavirus, #wuhanvirus, and #fuck_coronavirus.

## 5 CONCLUSION

This paper is aimed to analyze the impact of the novel coronavirus on the people by extracting and analyzing the sentiments of the top five most affected countries with the virus. We propose MACBiG-Net, which aims to extract the positive, negative, and neutral sentiment using word-level encoded attention and sentence level encoded attention. This is achieved by developing and labeling the COVID-19 Sentiment Dataset, which contains 4118 labeled tweets crawled from Twitter. Further, we visualize the sentiments of the people from January 31 to June 7 and summarize essential observations. We conclude that during the pandemic, lots of sentiments were being generated, and people were posting their views or opinions on several topics like frontline workers, upsurge in active cases, travel restrictions, and about the virus itself. Much fake news was also generated during this time, which was misleading the people. However, despite being stressed, people appreciated the efforts of all frontline workers and motivated each other to follow all precautions.

---

[10] https://timesofindia.indiatimes.com/blogs/toi-editorials/social-media-menace-beware-of-fake-news-going-viral-faster-than-covid-19-it-will-cost-lives-literally/

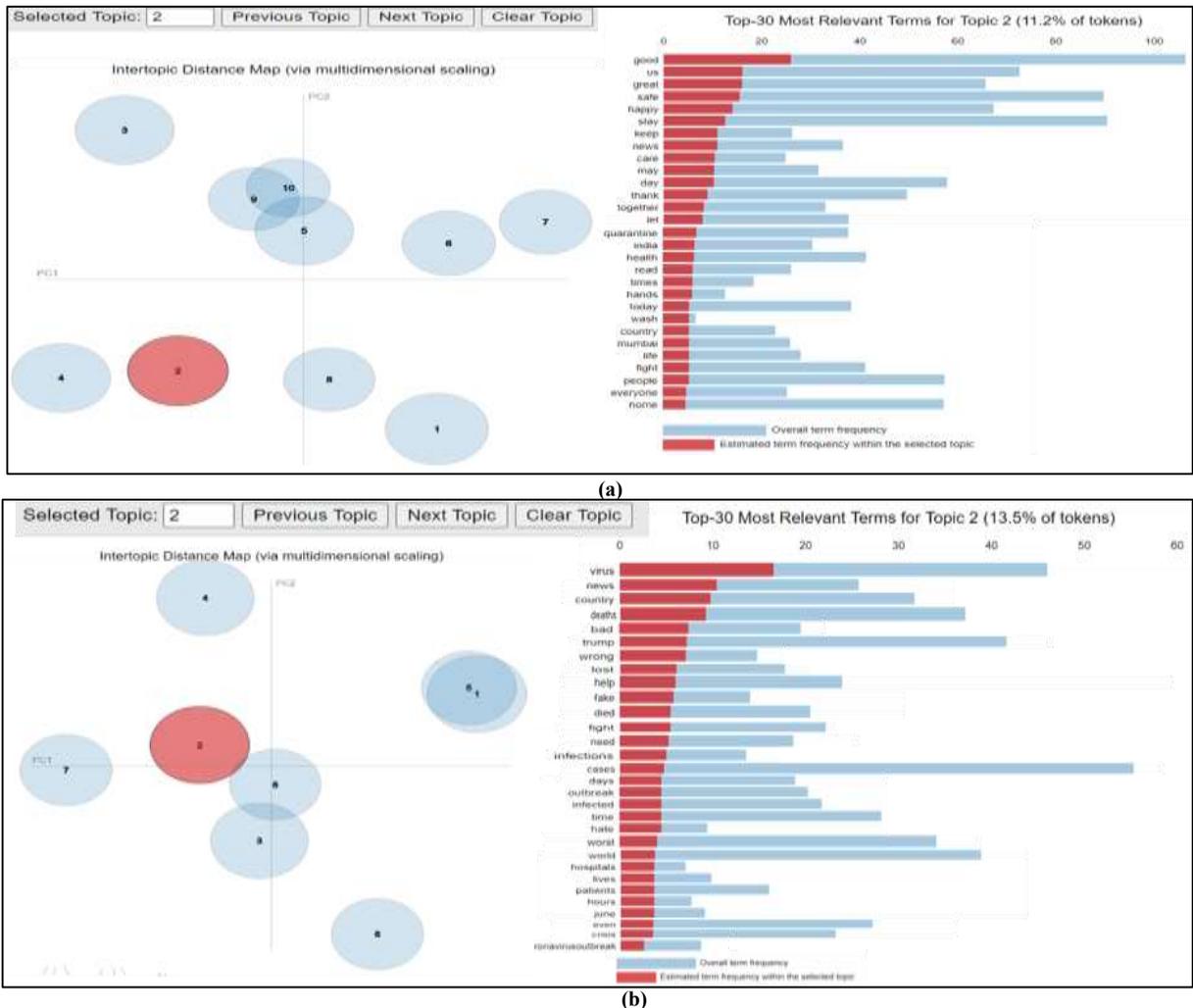

**Fig. 8. Identifying the hidden information in document through LDAvis visualization for (a) Positive and (b) Negative sentiment**

In the future, we aim to expand our study to analyze the impact of the virus on the rest of the world by doing in-depth research on several strategies like home-quarantine, work from home, social distancing undertaken to fight the deadly virus. We also aim to study the sentiments generated by other popular nations in the world.

## REFERENCES

[1] R. M. Anderson, H. Heesterbeek, D. Klinkenberg and T. D. Hollingsworth, "How will country-based mitigation measures influence the course of the COVID-19 epidemic?," *The Lancet,* vol. 395, no. 10228, pp. 931-934, 2020.

[2] J. Hellewell, S. Abbott, A. Gimma, N. I. Bosse, C. I. Jarvis, T. W. Russell, J. D. Munday, A. J. Kucharski and W. J. Edmunds, "Feasibility of controlling COVID-19 outbreaks by isolation of cases and contacts," *The Lancet Global Health,* vol. 8, no. 4, pp. e488-e496, 2020.

[3] A. Yadav and D. K. Vishwakarma, "A comparative study on bio-inspired algorithms for sentiment analysis," *Cluster Computing,* p. 1–21, 2020.

[4] M. Tsytsarau and T. Palpanas, "Managing Diverse Sentiments at Large Scale," *IEEE Transactions on Knowledge and Data Engineering,* vol. 28, no. 11, pp. 3028-3040, 2016.

[5] S. M. Mohammad, P. Sobhani and S. Kiritchenko, "Stance and sentiment in tweets," *ACM Transactions on Internet Technology (TOIT),* vol. 17, no. 3, pp. 1-23, 2017.

[6] W. Zhao, Z. Guan, L. Chen, X. He, D. Cai, B. Wang and Q. Wang, "Weakly-supervised deep embedding for product review sentiment analysis," *IEEE Transactions on Knowledge and Data Engineering,* vol. 30, no. 1, pp. 185-197, 2017.

[7] T.-c. Huang , R. N. Zaeem and K. S. Barber, "It Is an Equal Failing to Trust Everybody and to Trust Nobody: Stock Price Prediction Using Trust Filters and Enhanced User Sentiment on Twitter," *ACM Transactions on Internet Technology (TOIT),* vol. 19, no. 4, pp. 1-20, 2019.

[8] C. Huang, W. Jiang, J. Wu and G. Wang, "Personalized Review Recommendation based on Users' Aspect Sentiment," *ACM Transactions on Internet Technology (TOIT),* vol. 20, no. 4, pp. 1-26, 2020.

[9] A. Yadav and D. K. Vishwakarma, "A deep learning architecture of RA-DLNet for visual sentiment analysis," *Multimedia Systems,* vol. 26, p. 431–451, 2020.

[10] K. Denecke and Y. Deng, "Sentiment analysis in medical settings: New opportunities and challenges," *Artificial Intelligence in Medicine,* vol. 64, pp. 17-27, 2015.

[11] I. Korkontzelos, A. Nikfarjam, M. Shardlow, A. Sarker, S. Ananiadou and G. H. Gonzalez, "Analysis of the effect of sentiment analysis on extracting adverse drug reactions from tweets and forum posts," *Journal of Biomedical Informatics,* vol. 62, pp. 148-158, 2016.

12
12
12
12

[12] A. Yadav and D. K. Vishwakarma, "Sentiment analysis using deep learning architectures: a review," *Artificial Intelligence Review,* vol. 53, no. 6, pp. 4335-4385, 2019.

[13] Y. Wang, Q. Chen, M. Ahmed, Z. Li, W. Pan and H. Liu, "Joint Inference for Aspect-level Sentiment Analysis by Deep Neural Networks and Linguistic Hints," *IEEE Transactions on Knowledge and Data Engineering,* pp. 1-12, 2019.

[14] F.-C. Yang, A. J. Lee and &. S.-C. Kuo, "Mining Health Social Media with Sentiment Analysis," *Journal of medical systems*, vol. 40, no. 11, pp. 1-8, 2016.

[15] S. Sabra, K. M. Malik and M. Alobaidi, "Prediction of venous thromboembolism using semantic and sentiment analyses of clinical narratives," *Computers in Biology and Medicine,* vol. 94, pp. 1-10, 2018.

[16] M. Moh, T. Moh, . Y. Peng and L. Wu, "On adverse drug event extractions using twitter sentiment analysis," *Network Modeling Analysis in Health Informatics and Bioinformatics,* vol. 6, no. 1, pp. 1-12, 2017.

[17] S. M. Jiménez-Zafra, M. T. Martín-Valdivia, M. D. Molina-González and L. A. Ureña-López, "How do we talk about doctors and drugs? Sentiment analysis in forums expressing opinions for medical domain," *Artificial intelligence in medicine,* vol. 93, pp. 50-57, 2019.

[18] N. Limsopatham and N. Collier, "Normalising Medical Concepts in Social Media Texts by Learning Semantic Representation," *Proceedings of the 54th Annual Meeting of the Association for Computational Linguistics,* p. 1014–1023, 2016.

[19] Y. Chen, B. Zhou, W. Zhang, G. Gong and G. Sun, "Sentiment Analysis Based on Deep Learning and Its Application in Screening for Perinatal Depression," *IEEE Third International Conference on Data Science in Cyberspace,* pp. 451-456, 2018.

[20] H. Grisstte and E. Nfaoui, "Daily life patients Sentiment Analysis model based on well-encoded embedding vocabulary for related-medication text," *IEEE/ACM International Conference on Advances in Social Networks Analysis and Mining,* pp. 921-928, 2019.

[21] H. Talpada, M. N. Halgamuge and N. T. Q. Vinh, "An Analysis on Use of Deep Learning and Lexical-Semantic Based Sentiment Analysis Method on Twitter Data to Understand the Demographic Trend of Telemedicine," *11th International Conference on Knowledge and Systems Engineering (KSE), Vietnam,* pp. 1-9, 2019.

[22] Z. Yang, D. Yang, C. Dyer, X. He, A. Smola and E. Hovy, "Hierarchical attention networks for document classification," *Proceedings of the 2016 conference of the North American chapter of the association for computational linguistics: human language technologies, San Diego,* pp. 1480-1489, 2016.

[23] S. Hochreiter and J. Schmidhuber, "Long short-term memory," *Neural computation,* vol. 9, no. 8, pp. 1735-1780, 1997.

[24] A. Graves and J. Schmidhuber, "Framewise phoneme classification with bidirectional LSTM and other neural network architectures," *Neural networks,* vol. 18, no. 5-6, pp. 602-610, 2005.

[25] Y. Kim, "Convolutional neural networks for sentence classification," *Empirical Methods in Natural Language Processing,* pp. 1746-1751, 2014.

[26] S. Lai, L. Xu, K. Liu and J. Zhao, "Recurrent convolutional neural networks for text classification," *Twenty-ninth AAAI conference on artificial intelligence,* p. 2267–2273, 2015.

[27] Z. Lin, M. Feng, C. Nogueira, M. Yu, B. Xiang, B. Zhou and Y. Bengio, "A structured self-attentive sentence embedding," *International Conference on Learning Representations,* 2017.

[28] Y. Wang, A. Sun, J. Han, Y. Liu and X. Zhu, "Sentiment Analysis by Capsules," *Proceedings of the 2018 world wide web conference,* pp. 1165-1174, 2018.

[29] J. Devlin, M.-W. Chang, K. Lee and K. Toutanova, "BERT: Pre-training of Deep Bidirectional Transformers for Language Understanding," *Annual Conference of the North American Chapter of the Association for Computational Linguistics: Human Language Technologies (NAACL-HLT),* p. 4171–4186, 2019.

[30] C. Sievert and K. S. Shirley, "LDAvis: A method for visualizing and interpreting topics," *Proceedings of the workshop on interactive language learning, visualization, and interfaces,* pp. 63-70, 2014.

[31] K. Stevens, P. Kegelmeyer, D. Andrzejewski and D. Buttler, "Exploring topic coherence over many models and many topics," *Proceedings of the 2012 Joint Conference on Empirical Methods in Natural Language Processing and Computational Natural Language Learning,* pp. 952-961, 2012.

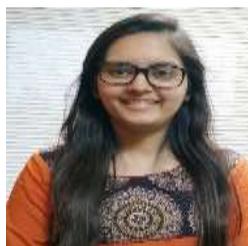

**Ashima Yadav** received B.Sc.(Hons) in Computer Science from University of Delhi, New Delhi, India in 2013, and M.C.A. from Guru Gobind Singh Indraprastha University, New Delhi, India in the year 2016. She is currently working towards the Ph.D. degree from the Department of Information Technology, Delhi Technological University, New Delhi, India. Her current research interest includes deep learning, natural language processing, machine learning, Emotion Recognition, and sentiment analysis.

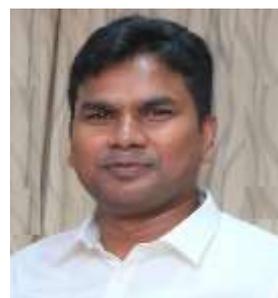

**Dinesh Kumar Vishwakarma (M'16, SM'19)** received the B.Tech. degree from Dr. Ram Manohar Lohia Avadh University, Faizabad, India, in 2002, the M.Tech. degree from the Motilal Nehru National Institute of Technology, Allahabad, India, in 2005, and the Ph.D. degree degree in the field of Computer Vision and Machine Learning from Delhi Technological University University (Formerly Delhi College of Engineering), New Delhi, India, in 2016. He is currently an Associate Professor with the Department of Information Technology, Delhi Technological University, New Delhi. His current research interests include Computer Vision, Machine Learning, Deep Learning, Sentiment Analysis, Fake News and Rumour Analysis, Crowd Behaviour Analysis, Person Re-Identification, Human Action and Activity Recognition. He is a reviewer of various Journals/Transactions of IEEE, Elsevier, and Springer. He has been awarded with "Research Excellence Award" by Delhi Technological University, Delhi, India in 2018, 2019 and 2020.